\documentclass[]{IEEEtran}

\usepackage{multicol}
\usepackage[bookmarks=true]{hyperref}
\hypersetup{
  linkcolor=blue
}

\usepackage{CJKutf8}
\usepackage{color}
\usepackage{enumerate}
\usepackage[ruled, linesnumbered]{algorithm2e}
\usepackage{amssymb, amsmath,mathtools}
\usepackage{stmaryrd}
\usepackage{graphicx}
\usepackage{setspace}
\usepackage{mathrsfs}
\usepackage{bm}
\usepackage[english]{babel}
\usepackage{blindtext}
\usepackage{epstopdf}

\newcommand{\bfxi}{\boldsymbol{\xi}}
\newcommand{\bfphi}{\boldsymbol{\phi}}

\newcommand{\bfb}{\bm{b}}
\newcommand{\bfc}{\bm{c}}

\newcommand{\bfe}{\bm{e}}
\newcommand{\bff}{\bm{f}}

\newcommand{\bfo}{\bm{o}}

\newcommand{\bft}{\bm{t}}
\newcommand{\bfu}{\bm{u}}

\newcommand{\bfy}{\bm{y}}

\newcommand{\bbR}{\mathbb{R}}

\newcommand{\calO}{{\cal O}}

\def\ie{i.e.\ }
\def\eg{e.g.\ }

\newcommand{\tf}[3]{\prescript{}{#1}{#2}^{#3}}
\newcommand{\bmT}{\bm{T}}
\newcommand{\bmR}{\bm{R}}
\newcommand{\bmX}{\bm{X}}

\newcommand{\bmA}{\bm{A}}
\newcommand{\bmB}{\bm{B}}

\newcommand{\SE}{\bm{S}\bm{E}(3)}
\newcommand{\bmSE}[1]{\bm{S}\bm{E}(#1)}
\newcommand{\SO}{\bm{S}\bm{O}(3)}

\newcommand{\bfSigma}{{\bf{\Sigma{}}}}

\begin{document}
\title{A probabilistic framework for tracking uncertainties in
  robotic manipulation}

\author{Huy Nguyen and Quang-Cuong Pham\thanks{The authors are with
    the School of Mechanical and Aerospace Engineering, Nanyang
    Technological University, Singapore. Corresponding author:
    Huy Nguyen (email: huy.nguyendinh09@gmail.com).} }
\maketitle

\begin{abstract}
  Precisely tracking uncertainties is crucial for robots to
  successfully and safely operate in unstructured and dynamic
  environments. We present a probabilistic framework to precisely keep
  track of uncertainties throughout the entire manipulation
  process. In agreement with common manipulation pipelines, we
  decompose the process into two subsequent stages, namely perception
  and physical interaction. Each stage is associated with different
  sources and types of uncertainties, requiring different
  techniques. We discuss which representation of uncertainties is the
  most appropriate for each stage (\eg as probability distributions
  in $\SE$ during perception, as weighted particles during physical
  interactions), how to convert from one representation to another,
  and how to initialize or update the uncertainties at each step of the
  process (camera calibration, image processing, pushing, grasping,
  etc.). Finally, we demonstrate the benefit of this fine-grained
  knowledge of uncertainties in an actual assembly task.
\end{abstract}


\section{Introduction}
\label{sec:introduction}

While tracking uncertainties has long been a central theme in mobile
robotics~\cite{thrun2005book}, it has received comparatively less
attention in industrial robotics in general, and robotic manipulation
in particular. Yet, precisely tracking uncertainties is crucial for
robots to successfully and safely operate in unstructured and dynamic
environments which, according to many reports\,\footnote{See for
  instance ``The Robotics Revolution: The Next Great Leap in
  Manufacturing'' by the Boston Consulting Group
  \url{https://www.bcg.com/publications/2015/lean-manufacturing-innovation-robotics-revolution-next-great-leap-manufacturing.aspx}.},
are becoming more common in the industry.

\begin{figure}[t]
  \centering
  \includegraphics[width=0.4\textwidth]{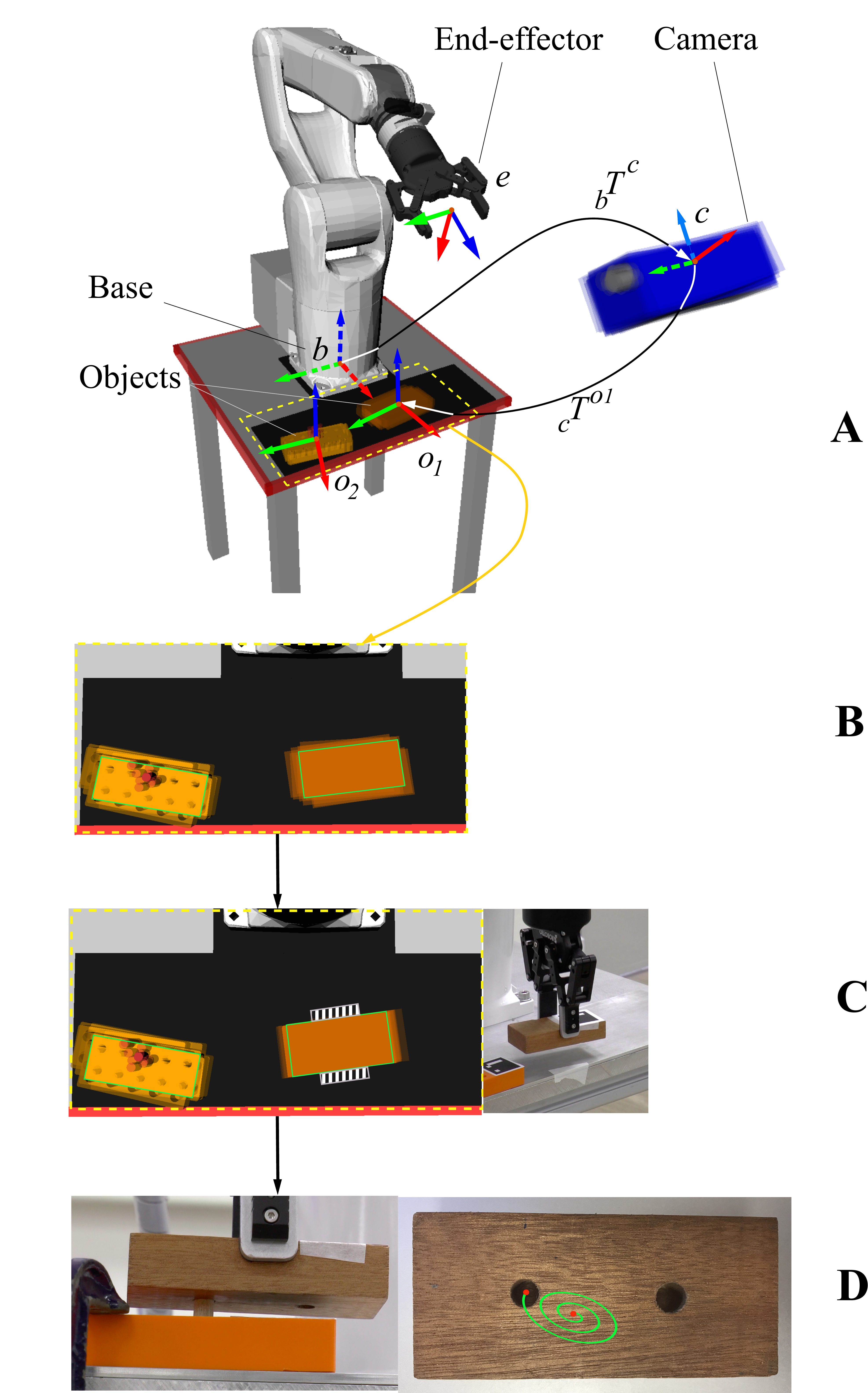}
  \caption{A typical assembly task. \textbf{A}: Setup, comprising a
    6-DOF robot equipped with parallel-jaw gripper, a 3D camera, and
    objects to be assembled. \textbf{B}: Uncertainty in object pose
    estimation. \textbf{C}: The uncertainty on the pose of the brown
    object has evolved under the grasping action. \textbf{D}: The
    elliptical search pattern for assembly is optimized based on the
    fine-grained knowledge of the uncertainties.}
  \label{fig:pininsertion}
\end{figure}

To illustrate, consider the assembly task depicted in
Fig.~\ref{fig:pininsertion}. For the robot to successfully assemble
the two objects, the poses of the objects relative to the robot must
be known to the system. However, uncertainties are introduced into the
object pose estimation during camera calibration and image
processing. Next, these uncertainties evolve during the physical
interactions between the robot and the objects (pushing, grasping,
etc.). Thus, the actual object poses before the last step of the
assembly might significantly differ from the assumed poses, which in
turn may cause assembly failures.

While it is not possible to totally eliminate all uncertainties --
because of the inherent noise introduced during the various perception
and action stages -- we argue in this paper that presicely keeping
track of the uncertainties can significantly improve the speed and
success rate of manipulation.


More specifically, we present a probabilistic framework to precisely
keep track of uncertainties throughout the entire manipulation
process. In agreement with common manipulation
pipelines~\cite{suarez2018can}, we decompose the process into two
subsequent stages, namely perception and physical interaction. Each
stage is associated with different sources and types of uncertainties,
requiring different techniques. We discuss which representation of
uncertainties is the most appropriate for each stage (\eg as
probability distributions in $\SE$ during perception, as weighted
particles during physical interactions), how to convert from one
representation to another, and how to initiate or update the
uncertainties at each step of the process (camera calibration, image
processing, pushing, grasping, etc.). Finally, we demonstrate the
benefit of this fine-grained knowledge of uncertainties in an actual
assembly task.


The remainder of the paper is organized as follows. Section~\ref{sec:related} sumarizes related works. In Section~\ref{sec:framework}, we present our framework for
precisely tracking uncertainties in manipulation. In
Section~\ref{sec:experiment}, we demonstrate the benefit of the
framework in several physical experiments. Finally, in
Section~\ref{sec:conclusion}, we conclude and sketch some directions
for future research.

\section{Related works}
\label{sec:related}
\subsection{Uncertainty in robotics}

Some fundamental works on modelling spatial uncertainties were studied
in \cite{smith1990ARV}, \cite{durrant1998JRA}, \cite{Wang08IJRR}, \cite{Barfoot14tr}. The above developments have been utilized to
establish a theoretical basis for a popular subfield of the
\emph{simultaneous localization and mapping} (SLAM) problem
\cite{thrun2005book,durrant2006RAM}. However, most existing researches
have only focused on \emph{mobile robotics}. Not much attention was paid explicitly to the modelling, estimating and
tracking the uncertainties in \emph{manipulation tasks}. In fact, most
manipulation tasks are sequences of primitive phases
consisting of calibration, perception and manipulating
actions. Existing works, however, have only focused on estimating
uncertainties in individual phases, not throughout the whole process,
where multiple sources of uncertainties interact and compound~\cite{sallinen2003modelling}.

The works closest to our paper are \cite{Su91icra,su1992manipulation},
in which authors presented a methodology for manipulating and
propagating spatial uncertainties in a robotic assembly system for
generating executable actions for accomplishing a desired
task. Although the works proposed a novel framework to work with
uncertainties, there was no discussion on how these uncertainties are
estimated. Based on those works, \cite{sallinen2003modelling} proposed
methods for estimating the geometrical relationships between
coordinate frames and the spatial uncertainties in estimated model
parameters. However, later phases (\ie manipulating actions) where
contacts between robot(s) and the environment occur have not been
investigated. On another note, their estimation model were based on
an Euler angles parameterization, as opposed to the $\SE$ formulation,
is well-known to involve singularities.

\subsection{Representation and propagation of uncertainty}

In early works by \cite{Taylor1976thesis} and \cite{brooks1982IJRR},
the uncertainty of a pose is simply represented by worst-case bounds which
include all possible errors (min-max approach).This simple approach usually results in conservative estimates which make it
difficult to apply in decision-making process.

To address this limitation, the probabilistic approach which uses the
calculus of probability theory and assign probabilities to all
potential positions of the object has been proposed. The pose of
an object is now represented by a probability distribution over the
space \cite{Su91icra}, \cite{su1992TSMC}. As a result, the
probabilistic representation can make use of probability theory and,
thus, provide more uncertainty manipulation methodologies
(\eg propagation, fusion, etc.) \cite{smith1990ARV}, \cite{durrant1998JRA}, \cite{brooks1985icra}, \cite{smith1986jra}. Recently, this probabilistic approach has been
further investigated in \cite{Wang08IJRR}, \cite{Barfoot14tr}, \cite{chirikjian2009vol1},
\cite{chirikjian2011vol2} where they provided a
rigorous treatment of representing and propagating uncertainty on
$\SO$ and
$\SE$. 

\subsection{Alternative approaches to deal with uncertainty in manipulation}

A common approach to deal with uncertainties while performing
manipulation tasks is to plan the geometric motions of the objects and
then execute these motions on a compliant robot. This simple approach
has proved to be a key component for the success of many manipulation
tasks
\cite{suarez2018can,knepper2013ikeabot,wahrburg2014contact,van2018comparative}. To
compute robust motions under the presence of motion and sensing
uncertainties, many works have incorporated such information into
their
planners~\cite{phillips2017planning,sieverling2017interleaving,wirnshofer2018robust}. This
  line of research are generally referred to as belief space planning
  \cite{platt2010belief,melchior2007particle,hauser2010randomized}. Even
though such approaches have enabled many works to reliably accomplish
complex manipulation tasks in realistic, uncertain environment, they
commonly requires a set of hypotheses of pose for every manipulated
object in the environment. In fact, these sets of hypotheses are often
assumed to be much larger than the real ones to account for ambiguous
estimations. Consequently, robots have to perform many more redundant
motions and spend more time to reason the effects of these
uncertainties. This lack of access to explicit uncertainty information
motivates us to study and proposed a new probabilistic framework for
tracking uncertainties in manipulation tasks. In turn, the
capabilities to obtain such fine-gain information enabled by our
framework can also be utilized in these planning approaches to achieve
a higher level of performance when dealing with uncertainties.

\section{A probabilistic framework for tracking uncertainty}
\label{sec:framework}
In general, most typical manipulation tasks can be decomposed into
two subsequent stages, namely perception and physical interactions,
see Figure~\ref{fig:pininsertion}. Each stage is associated with
different sources and types of uncertainties, hence requiring
different techniques to deal with. The rest
of this section will discuss which representation of
uncertainties is the most appropriate for each stage, and how to
initialize or update the uncertainties at each step of both stages.

\subsection{Choices of uncertainty representations}
\subsubsection{During perception stage}
During this stage, we choose to represent uncertainties of
3D poses as probability distribution in $\SE$. This is
accomplished by storing the means as a (singularity-free)
transformation matrix and using a (constraint-sensitive) perturbation
of the pose (with associated covariance matrices). In particular, we model
the uncertainties of the rotation and the translation parts separately
as follows:

We assume that the rotation part of a pose is corrupted
as below
\begin{eqnarray}
\label{eq:rotationnoise}
\bmR = \exp([\bfxi_{\bmR}])\bar{\bmR},
\end{eqnarray}
where $\bar{\bmR} \in \SO$ is the
mean of $\bmR$, and the small perturbation variable
$\bfxi_{\bmR}\in\bbR^3$ is zero-mean
Gaussian with covariance matrice
$\bfSigma_{\bmR}$.

The translation part of a pose is corrupted as below
\begin{eqnarray}
\label{eq:translationnoise}
\bft = \bfxi_{\bft} + \bar{\bft}, 
\end{eqnarray} where $\bar{\bft}\in\bbR^3$ is the mean of $\bft$, and
$\bfxi_{\bft} \in\bbR^3$ is the zero-mean Gaussian perturbation with
covariance matrice $\bfSigma_{\bft}$.

This represenation is free of singularities and avoids
the need to enforce constraints when
solving optimal estimation problems. Moreover, such
representation also allows us to utilize the Lie
groups and their associated mathematical machinery to provide a rigorous
treatment to address many essential operations, \ie propagation and
fusion of uncertainties (see later in \ref{subsec:perception}). Thanks
to these advancement, we are able to store uncertainty
information as analytical form and evaluate the uncertainties of
object pose estimation in timely fashion. 

In fact, the representation also implies that rotation and translation noises
are independent\footnote{Since there is in
general no bi-invariant distance on $\SE$~\cite{PR97acm},
when solving optimal estimation problems, finding the rotation and
translation components simultaneously would require a non-trivial
rotation/translation weighting in any cases. As a result, we choose to
model the uncertainties of the rotation and the translation parts
separately instead of working directly on $\SE$.}. Because of this assumption, the space we consider is,
strictly speaking, $\SO \times \bbR^3$, rather than
$\SE$.

Under this representation, belief states are represented by multivariate normal distributions (Gaussian
distributions). Although it can bring us a fast, analytical method to
estimate uncertainties in the perception stage, there are a number of
shortcomings. Most importantly, Gaussian distributions are unimodal
(probability density exhibits single peak). However, in other cases
where one must handle multimodal distributions, \ie during the physical
interaction stage, particles are more preferable.

\subsubsection{During physical interaction stage}
Unlike the perception stage where camera measurements are
easily used to infer the poses of the objects, physical interactions only
provide local information about those poses. Consequently, motion and
observation models during this stage are often highly non-linear and lack of analytic
derivatives. Moreover, the resulting belief state is usually non-Gaussian and may be
multi-modal. This, therefore, naturally leads to the use of particle filter and its variants to
track the belief state during the physical interaction stage.

\subsection{Perception}
\label{subsec:perception}
Regarding the perception stage, we are interested in estimating the
poses of the objects and their associated uncertainties with respect
to the robot base frame. As illustrated in
Figure~\ref{fig:pininsertion}, such uncertainties are
definitely originated from three main sources which include: (i) the
uncertainty of the object pose estimation in the camera frame, (ii)
the uncertainty of the camera position (commonly known as the hand-eye
transformation), \emph{and} (iii) the uncertainty of the robot
end-effector positioning.

\begin{equation}
\tf{\bfb}{\bmT}{\bfo 1} = \tf{\bfb}{\bmT}{\bfc}\tf{\bfc}{\bmT}{\bfo 1}
\end{equation}

We note that in this setting the camera is mounted at a fixed position
in the environment (not on the robot end-effector). Hence, the
uncertainty of the robot end-effector positioning does not contribute
directly to the uncertainties of the object poses. Instead, it affects
the uncertainty of the camera position via the hand-eye calibration
process \cite{huy18tro}.

In this section, we shall provide a methodology to estimate explicitly
these sources of uncertainties. We will also give a
discussion on how the uncertainties can be propagated so that the
poses of the objects and their associated uncertainties with respect
to the robot base frame can be eventually obtained.

\subsubsection{Estimating uncertainties}

\begin{itemize}
\item \emph{Uncertainty on the robot positioning $\tf{\bfb}{\bmT}{\bfe}$}:
Thanks to the advancement of calibration methods and measurement
equipment, robots are now better built with higher accuracy. After calibration, a common
positioned control industrial robot can achieve sub-milimeter in the
mean position errors of the end-effector. For example, our experiments
show that the mean of the position errors of our calibrated manipulator is about
0.3~mm. This can be considered negligible compared to other sources
of uncertainties in our system, \eg the uncertainty of the pose
estimation using our camera system. Because of this result, we find it usually
safe to assume accurate joint measurements and robot model.

\item \emph{Uncertainty on the hand-eye calibration
    $\tf{\bfb}{\bmT}{\bfc}$}: Commonly, this problem can be formulated
  as: solve for $\bmX$ in $\bmA\bmX = \bmX\bmB$, where $\bmX$ is the
  unknown $4\times 4$ hand-eye transformation matrix and $\bmA$ and
  $\bmB$ are known $4\times 4$ transformation matrices.

  Here, we are interested, not merely in solving for the hand-eye
  transformation, but more comprehensively, in evaluating its
  covariance. To address this problem, we follow \cite{huy18tro} where
  they propose a novel algorithm to rigorously work out such a
  derivation. In particular, the technique exploits the benefits of
  applying optimization techniques directly on $\SE$ to obtain the
  derivation of the covariance of the hand-eye transformation.

\item \emph{Uncertainty on the camera-based pose estimation
    $\tf{\bfc}{\bmT}{\bfo}$}: Regarding the camera-based pose
  estimation problem, its uncertainty mainly depends on the intrinsic
  calibration of the camera. 
  However, a model to capture the uncertainties of the camera
  parameters is highly non-linear and difficult to be
  evaluated. Hence, we decide to estimate such uncertainty by drawing
  a number of samples and performing Monte Carlo estimation. Despite
  the fact that this approach is slow, it is proven to bring out a
  more accurate estimation. A more detailed clarification of how this
  Monte Carlo estimation performed in the real systems can be found in
  the Experimental Section of \cite{huy18tro}.
 \end{itemize}

\subsubsection{Propagating uncertainties}
\label{frameworkchap:propagation}

After obtaining all sources of uncertainties in the perception stage,
we are now in a position to estimate the uncertainty of the object pose with respect to the robot
base frame. 
In this work,
we chose to follow the method proposed
in~\cite{huy18tro} (Appendix A), where the covariance propagation
method are derived for the
case where rotation and translation are decoupled.

\subsection{Physical interactions}
\label{chapframeworksubsec:physicalinteractions}

In this work, we focus on manipulating tasks that usually perform on
industrial assembly where common elements, \ie force/torque sensors and
parallel-jaw grippers, are ubiquitous due to their strength,
robustness, cost-effectiveness, ease of integration, and many
more. The work in this paper will only cover two manipulating actions
commonly performed in manipulation, which are plannar grasping actions
and touch-based localization.

\subsubsection{Problem formulation}

Given the initial distribution of the objects obtained from the
previous perception stage, the goal is to update the object
distributions after the plannar grasping action or/and the touch-based
localization. As commonly known, the Bayesian filter is considered as
the most general algorithm for filtering a belief state given initial
knowledge and a sequence of actions and observations. Hence, the
problem will be cast into the Bayesian framework and be addressed as a
nonlinear filtering problem as shown below.

Let $\bmX$ be the state of a dynamical system which evolves under
actions $\bfu$ and provides observations $\bfy$. Starting with $P(\bmX_{0})$ -- the \emph{prior} distribution over the state
$\bmX$ -- the goal is to recursively update the following
conditional probability
\begin{equation}
\label{eqn:bayesupdate}
P(\bmX_{t+1}|\bfy) = \eta P(\bfy|\bmX_t) P(\bmX_t).
\end{equation}
Here $P(\bmX_{t+1}|\bfy)$ is known as the \emph{posterior}, which
represent our uncertain belief about the state $\bmX$ after having
incorporated the measurement $\bfy$. On the right-hand side, the first
factor $P(\bfy|\bmX_t)$ is the \emph{measurement probability},
which encodes the likelihood of the measurement given the state
(\emph{measurement model}). The second factor $P(\bmX_t)$ is the
\emph{prior}, which represents our belief about $\bmX$ before obtaining
the measurements $\bfy$. The factor $\eta$ is a normalizing factor
independent of the state $\bmX_t$ and needs not be computed.



Over the next sections, we will discuss how to apply this filter into two
common contact actions, which are plannar grasping actions
and touch-based localization.

\subsubsection{Plannar grasping}
\label{subsec:plannar}

Regarding plannar grasping action, the state is the pose $\bmX \in \bmSE{2}$ of the
manipulated object. Actions are motions of the hand, given by the
velocity $\bfu$. During contact, the object moves with
a velocity $\bff_{\bfphi} (\bmX, \bfu)$ where the function $\bff$
encodes the physics of the object motion in response to the intended motion of
the gripper. The parameter $\bfphi$ includes environmental properties. 
In this work, we particularly build analytical state
estimators to track the poses of the objects from the post-grasp gripper
distances based on the works from
\cite{mason1986mechanics, goyal1991planar}. This motion model is
constructed following the assumption of quasi-static rigid
body mechanics with Coulomb friction. Such assumption not only allows
us to attain more approachable and simpler models, but also well suit the scale and speed of our
application.

Relating to the Bayesian updates, the particle filter first
samples the particles ${\bmX^{i}}$ from the prior distribution, then
uses the motion model in order to \underline{sim}ulate and obtain
${\bmX^i}_{\mathrm{sim}}$. Once the final object poses are acquired, their
associated finger widths $d_{{\bmX^i}_{\mathrm{sim}}}$ can be estimated. This information will
be used to compute an important weight for each forward-simulated
particle. During the weighting step, the particles which are consistent
with our measurement $\bfy$ will be assigned with higher
probability. In particular, the post grasp distance measurements are
assumed to be corrupted by Gaussian noise with the variance
$\sigma_d$. The measurement probability 
is computed as follows
\begin{equation}
 P(\bfy|{\bmX^i_{t}}) = \eta_{\bfy} \exp{\left(-\frac{1}{2}
     \frac{(\bfy-d_{{\bmX^i}_{\mathrm{sim}}})^2}{{\sigma_d}^2}\right)},
\end{equation}
where $\eta_{\bfy}$ is a constant and will be taken into account
during the normalization.

\begin{figure}[t]
  \centering
  \includegraphics[width=0.4\textwidth]{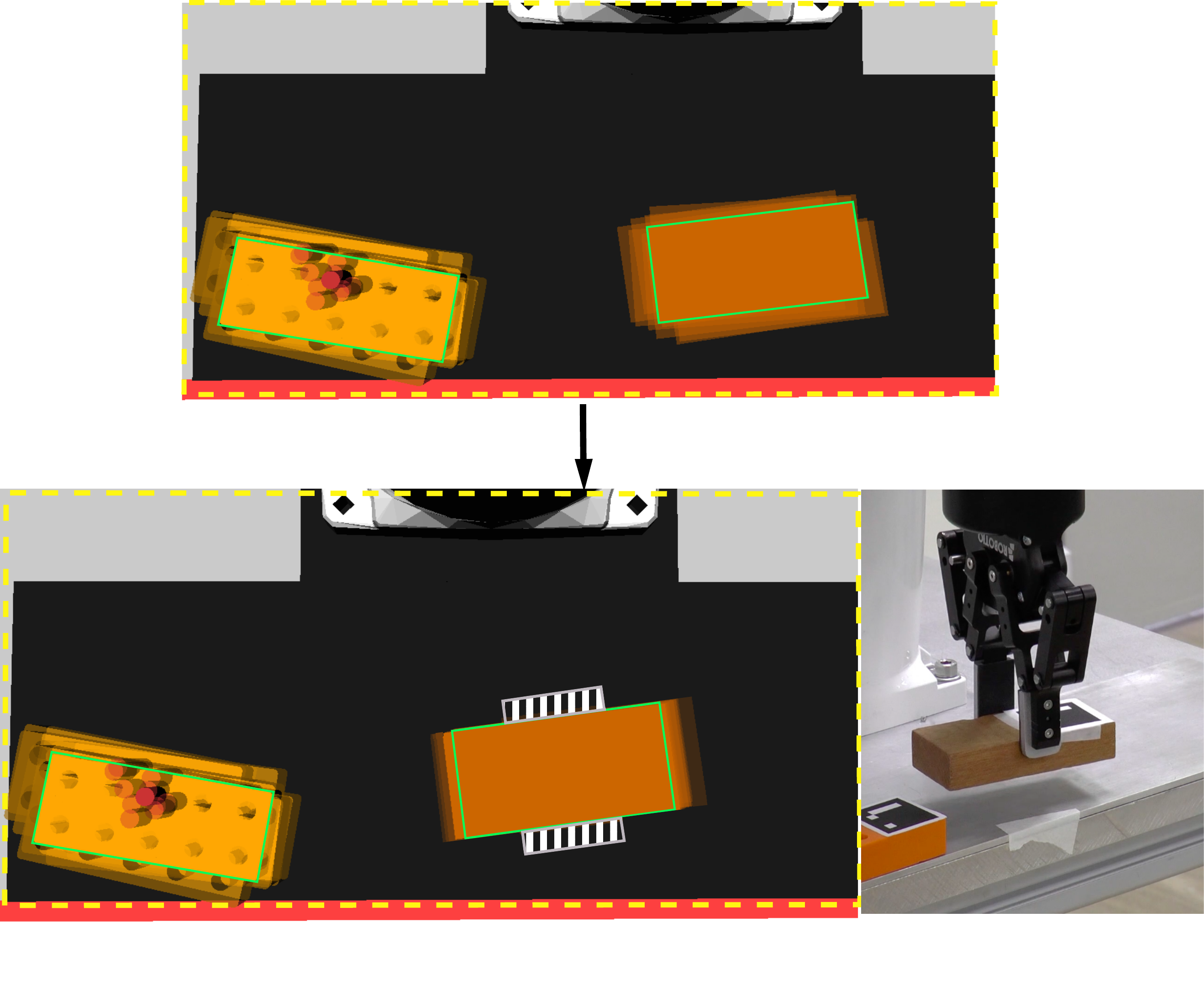}
  \caption{Once a plannar grasping action is
successfully performed, it sometimes can significantly improve the estimation of
the objects in the closing direction of the gripper.}
  \label{chapframeworkfig:plannargrasp}
\end{figure}

It is also worth-noting that once a plannar grasping action is
successfully performed, it can significantly improve the estimation of
the objects in the closing direction of the gripper. Especially, in the
cases where the geometries of the grasped objects are simple,
\eg rectangular boxes, the uncertainties of the objects
could shrink to an one-dimensional distribution, see
Figure~\ref{chapframeworkfig:plannargrasp}. Besides, in this
work, we also assume that the objects do not end up slipping out or being
jammed at an undesired position. This assumption is considered
reasonable in this case because the object geometries are simple and the slow movement of
the fingers is always under control.

\subsubsection{Touch-based localization}
\label{frameworkchapsubsec:touch}

With regards to touch-based localization, this type of interaction is
often employed to further improve the pose estimation of the
objects. As mentioned earlier, the objects to be located are, in
general, assumed to be static during the measurement collection. This
commonly-chosen assumption is realistic in such cases of handling very
slight contact and the objects that are heavy or mounted on a support
fixture to prevent their possible movements.


\begin{figure}[t]
  \centering
  \includegraphics[width=0.45\textwidth]{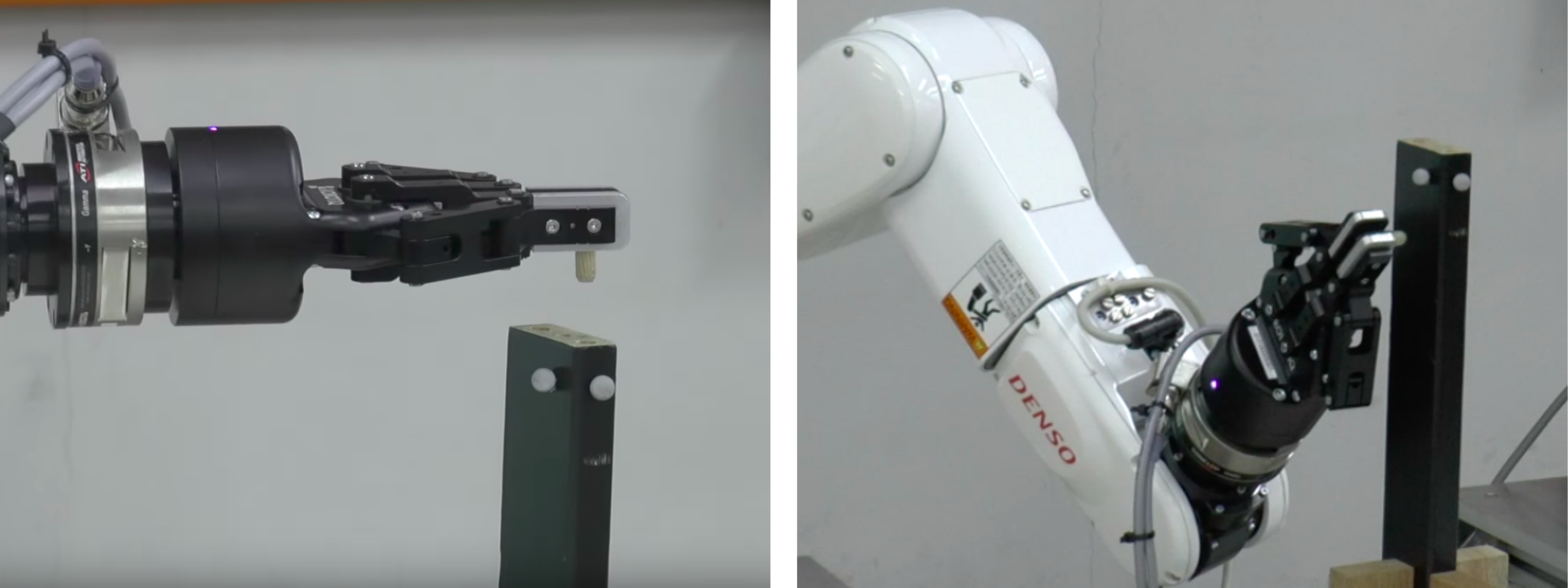}
  \caption{The object localization problem via touch. The
    photo shows the robot interacting with a wooden stick using a pin
    and its force/torque sensor.}
  \label{fig:touch-framework}
\end{figure}

In this problem, the state is the pose $\bmX \in \SE$ of an object
$\calO$ with a known shape. The measurements
$\bfy={\bfy_0,...,\bfy_n}$ are obtained by touching the object with
the end-effector of the robot (as shown in Figure
\ref{fig:touch-framework}). Each measurement
$\bfy_k := (\bfy^{\mathrm{pos}}_k,\bfy^{\mathrm{nor}}_k)$ consists of an contact
position $\bfy^{\mathrm{pos}}_k$ and a contact normal $\bfy^{\mathrm{nor}}_k$.

In fact, many works have been
proposed and are able to solve the 6-DOF localization problem
efficiently and reliably. In this work, we use the \emph{proximity
  measurement model} and \emph{Scaling Series} method in
\cite{petrovskaya2011global,nguyen2017touch} to perform the touch-based localization
owing to its computational efficiency.

\section{Application to part assembly}
\label{sec:experiment}

In this section, we discuss an application of our framework to parts
assembly tasks. The robotic platform used in this work is
characterized by cost-efficient, off-the-shelf components combined
with a classical position-control industrial
manipulator. In particular, the main components of our platform are:

\begin{itemize}
\item 1 Denso VS060: Six-axis industrial manipulator.
\item 1 Robotiq Gripper 2-Finger 85: Parallel adaptive gripper
  designed for industrial applications. Closure position, velocity and
  force can be controlled. The gripper opening goes from $0$ to $85~\mathrm{mm}$. The grip force ranges from $30$ to $100~\mathrm{N}$.
\item 1 ATI Gamma Force-Torque (F/T) Sensor: It measures all six
  components of force and torque. They are calibrated with the
  following sensing ranges: $\bff = [32, 32, 100]~\mathrm{N}$ and
  $\tau = [2.5, 2.5, 2.5]~\mathrm{Nm}$.
\item 1 Ensenso 3D camera N35-802-16-BL
\end{itemize}

\subsection{Single pin insertion}

To illustrate the effectiveness of the framework, we first consider a dexterous task: a cylindrical peg-in-hole task. In
this task, two rectangular boxes $(20\times50\times110)~\mathrm{mm}$ are placed
randomly on a known table. The first box $\texttt{Obj1}$, also the
static one, is clamped down on the table with a cylindrical pin
$(r=4~\mathrm{mm}, l=30~\mathrm{mm})$ pre-inserted. The second box-$\texttt{Obj2}$,
considered as the movable one, will be grasped by the robot. Besides,
the camera mounted on a fixed tri-pod will be used to localize the two
above boxes (see Figure~\ref{fig:pininsertion}). As shown in the
Figure, to perform its task, the robot arm needs to grasp
$\texttt{Obj2}$ and insert it into $\texttt{Obj1}$. After it completes
the insertion, it will then release $\texttt{Obj2}$. Note that the
geometries of all objects are assumed to be known.

This is a good example for our framework illustration since most of
the capabilities we have developed are required to perform this
task. The task also constitutes one of the key steps in many assembly
processes. In addition, it yields a fully quantifiable way to measure
the performance of a robotic manipulation system. For these reasons,
it is considered as a good test for the generalizability and
simplicity of our approach.


Regarding the execution of the task, the process is naturally divided
into three sequent sub-tasks as below:

\begin{itemize}
\item{locate the two objects by using the camera;}
\item{pick $\texttt{Obj2}$;}
\item{compliant insertion of the two objects.}
\end{itemize}

\subsubsection{Perception}
A camera mounted on a fixed tri-pod is
used to localize the two boxes. In this setting, we note that the
tranformation of an object relative to the robot base is given by
\begin{equation}
\label{eq:obj}
\tf{\bfb}{\bmT}{\bfo} = \tf{\bfb}{\bmT}{\bfc} \tf{\bfc}{\bmT}{\bfo}.
\end{equation}
This implies that the uncertainty of the robot positioning does not
directly affect the uncertainty of the object pose with respect to the
robot base. However, as mentioned in Section \ref{subsec:perception},
the uncertainty of the robot positioning (if existing) also affects
the estimation of the $\tf{\bfb}{\bmT}{\bfc}$ via the hand-eye
calibration. Again, it is noted that we assume the uncertainty of our
manipulator is negligible compared to the uncertainty of the pose
estimation using our camera system. We follow the proposed framework
in Section~\ref{subsec:perception} to estimate the poses of the two
objects and their associated uncertainties.

\subsubsection{Pick $\texttt{Obj2}$}
Once the two objects are located, the robot arm needs to perform the
picking task. In fact, this task resembles the plannar grasping action
which has been discussed earlier in Section \ref{subsec:plannar}.

To be specific, before grasping $\texttt{Obj2}$, the robot arm moves
its gripper to a pre-grasp position which is above the object. The
gripper, then, gradually close its fingers until it exceeds the force
limit. After the execution of the grasping action, we read the
distance between the fingers from the Robotiq gripper encoder, and
update the distribution of the object. In this step, the previous
information obtained from the perception sub-task is used as the
initial uncertainty region of $\texttt{Obj2}$. Note that since we
assume the table plane to be known, we then project $\texttt{Obj2}$
uncertainty region onto this surface. Moreover, the pose of
$\texttt{Obj2}$ and its associated uncertainty will be presented in
the gripper local frame as $\texttt{Obj2}$ is now rigidly grasped by
the gripper.

\subsubsection{Compliant pin insertion}
Next, pin
insertion is the final sub-task to be executed. Due to the
uncertainties on the position of the objects ($\texttt{Obj1}$ and
$\texttt{Obj2}$), the exact position of the hole/pin is
unknown. Moreover, given the parameters of the peg-in-hole set-up, we
observe that the insertion will fail in case the position errors are
more than $0.5~\mathrm{mm}$. To cope with such uncertainty, many researchers have
proposed to apply force-controlled exploration. For example,
\cite{suarez2018can} used a super-imposed spiral search
pattern. In their setting, the hole position is uncertain. The robot
is controlled to follow the spiral pattern while maintaining the
contact between the pin and the hole. Once the hole is found, the
exploration will be immediately terminated. Even though this strategy
is able to precisely and consistently detect holes and perform tight
pin insertions ($100\%$ success across the $28$ insertions), it is
indeed a ``blind searching'' algorithm since there is no information
about the uncertainties of the objects to be used. This limitation
motivates us to study how such uncertainties can be employed to design
a better compliant insertion strategy as below.

After the gripper grasps $\texttt{Obj2}$, the robot arm moves the
$\texttt{Obj2}$ to the pre-inserting position. It, then, moves down until it
touches the surface of $\texttt{Obj1}$. Once they are in contact, the relative
transformation between the two objects is as follows:
\begin{equation}
\label{eqn:obj2obj1}
\tf{\bfo 1}{\bmT}{\bfo 2} = \tf{\bfo
  1}{\bmT}{\bfb}\tf{\bfb}{\bmT}{\bfe}\tf{\bfe}{\bmT}{\bfo 2},
\end{equation}
where $\tf{\bfo1}{\bmT}{\bfb}$ is the transformation of the base with
respect to $\texttt{Obj1}$ frame, $\tf{\bfe}{\bmT}{\bfo 2}$ is the relative
tranformation between $\texttt{Obj2}$ and the end-effector
(gripper). As the uncertainty of $\tf{\bfe}{\bmT}{\bfo 2}$ is
represented by particles, we assume it to be a single peak Gaussian distribution and
perform an empirical estimation to obtain its parameters\footnote{Such
  assumption is suitable for our experiment where object geometry is
  simple. For better capture of the multi-model of the distribution, the Gaussian
  mixture model is recommended.}. The uncertainty of
$\tf{\bfo 1}{\bmT}{\bfo 2}$ is, then, obtained by the forward
propagation method. The search pattern is now optimized based on such
fine-gaind knowledge of the uncertainty. Instead of
using the ``circular'' spiral trajectory, we change the spiral pattern
which follows the elliptical shape of the one-standard-deviation
covariance ellipsoid of the $\tf{\bfo 1}{\bmT}{\bfo 2}$ distribution
(see Figure~\ref{fig:pininsertion}).

Compared to the traditional ``circular'' spiral search
strategy, our new proposed spiral pattern can successfully perform the
insertion at a faster speed. The Table \ref{tab:frametab} shows the
experimental results of over 50 insertions for each method. As
expected, we achieve $100\%$ success across all the insertions. In
addition, our method requires only $11.2\pm4.5$ seconds to insert
successfully, which is about two times faster compared to the
traditional search strategy.

\begin{table}[h]
 \centering
    \begin{tabular}{|c|c|c|}
      \hline
      &Circular spiral search&Elliptical spiral search\\
      \hline
      Time(s) &$22.8\pm5.7$ &$11.2\pm4.5$\\
      \hline
    \end{tabular}
    \caption{We compare our elliptical spiral search with the traditional
``circular'' spiral search strategy.}
\label{tab:frametab}
\end{table}

\subsection{Double pin insertion}
\begin{figure}[t]
  \centering
  \includegraphics[width=0.5\textwidth]{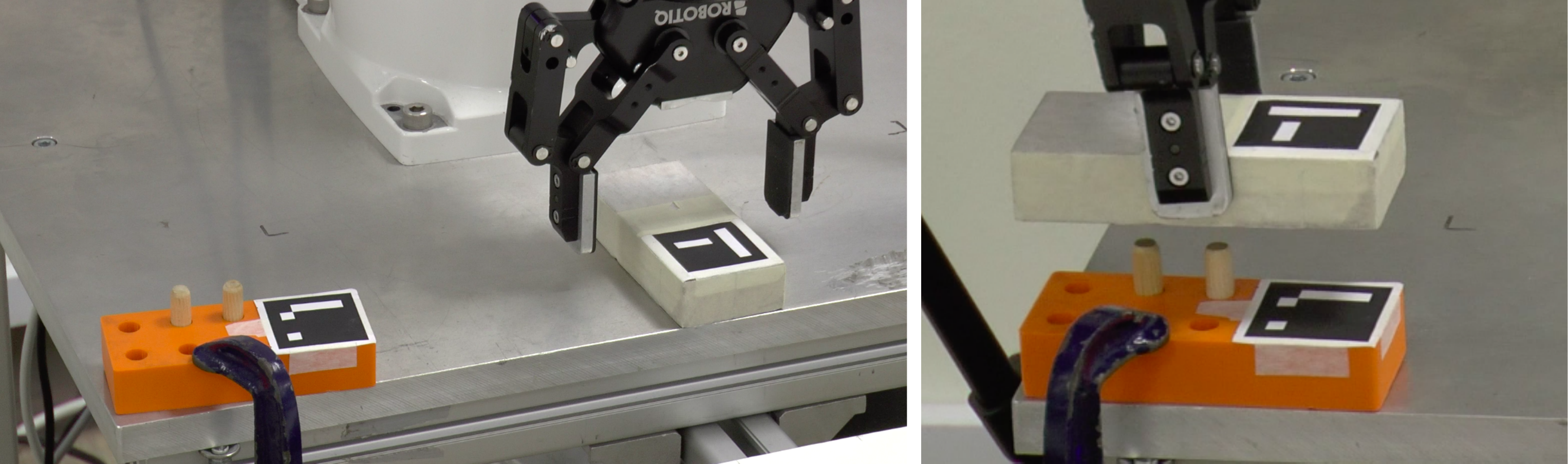}
  \caption{We exploit the benefits of having access to uncertainties
    information in a double pins insertion task.}
  \label{fig:task2}
\end{figure}

In previous experiment, we have studied \emph{single} pin insertion task in which only the uncertainties in
the translation parts of the object poses were employed to derive the
new search strategy. We now study another example where both
uncertainties of rotation and translation parts are used in planning
the insertion strategy. In particular, we keep the experiment setting
similar to last task except that both objects now have two pins/holes
(see Figure~\ref{fig:task2}).

In this experiment, we assume $\texttt{Obj2}$ is
fully constrained and its pose is determined before the execution of
the compliant double pin insertion. Note that this can be done by
simply bringing the object to contact with a known surface in order to
futher reduce the uncertainty after the grasping action.

To perform
the double pin insertion, we decouple the task into two
force-controlled explorations. First, the robot arm moves
$\texttt{Obj2}$ to the pre-inserting position, slightly tilts
$\texttt{Obj2}$, then starts moving down until contact is detected. We
then perform the elliptical spiral search to find the precise position
of the first pin/hole using the information of the uncertainty of the
translation part. Second, we find the next pin/hole by simply rotating
$\texttt{Obj2}$ around the first pin/hole axis while maintaining
contact with the surface. In this case, the information of the
uncertainty of the rotation part are employed to determine the bound
angles of the exploration. As expected, we also achieve $100\%$
success across 25 attempts. The average running time of the
force-controlled exploration is $28.3\pm7.1$ seconds.

The video of the two experiments can be found at \href{https://youtu.be/rbo4QL40fqU}{\texttt{https://youtu.be/rbo4QL40fqU}}.

\section{Conclusion}
\label{sec:conclusion}
In this paper, we have presented a probabilistic framework to
precisely keep track of the uncertainties throughout the entire
manipulation process. In order to do so, we decompose the
manipulation task into two subsequent stages, namely perception and
physical interactions. Each stage is associated with different sources
and types of uncertainties, requiring different techniques. We discuss
which representation of uncertainties is the most appropriate for each
stage (\eg as probability distributions in $\SE$ during perception, as
weighted particles during physical interactions), how to convert from
one representation to another, and how to initialize or update the
uncertainties at each step of the process (camera calibration, image
processing, pushing, grasping, etc.). 

We have shown that precisely keeping track of the uncertainties in the system can
significantly improve the speed and success rate of the
manipulation. For example, as shown in our
experiment on the single cylindrical peg-in-hole task, the spiral
search operation was accelerated by two times. We also demonstrated
how the information of uncertainties in the system can be applied into
more complex task, \ie double pin insertion. In fact, without such
information, it would be extremely challenging to perform the
mentioned tasks successfully.

In addition to the mentioned complex insertion tasks, we believe these
techniques could find application in various manipulation tasks that
require the knowledge of the uncertainties of the object poses. One application is to estimate the success probability of a particular action. This estimation plays an
important role as it enables the robot to improve the overall
success rate of the task by not performing unnecessary
motions. Another application is to plan the action that can actively
reduce the uncertainties in the assembly process and simultaneously
decrease the number of actions required. Moreover, future work can also inherit the proposed techniques to employ in bimanual
manipulation tasks with higher complexity.

It is also worth-noting that the benefits of the precise information
about the uncertainties inevitably comes with an expense. The reason
is that it takes more time for the robot to reason and obtain such
information from the measurements after each physical interaction. In
fact, both plannar grasping action and touch-based localization need
to be taken into account when dealing with this challenge. Regarding
the plannar grasping action, ~\cite{zhou2017probabilistic} have
proposed a solution in which the forward motion model is approximated
by kennel conditional density estimation (KCDE). Nevertheless, in
order to properly address this problem, further investigation to
improve the overall speed of the estimation process is required in
future works.

\section*{Acknowledgment}
This work was supported in part by NTUitive Gap Fund NGF-2016-01-028
and SMART Innovation Grant NG000074-ENG.

\bibliographystyle{IEEEtran}
\bibliography{framework.bbl}

\begin{thebibliography}{10}
\providecommand{\url}[1]{#1}
\csname url@samestyle\endcsname
\providecommand{\newblock}{\relax}
\providecommand{\bibinfo}[2]{#2}
\providecommand{\BIBentrySTDinterwordspacing}{\spaceskip=0pt\relax}
\providecommand{\BIBentryALTinterwordstretchfactor}{4}
\providecommand{\BIBentryALTinterwordspacing}{\spaceskip=\fontdimen2\font plus
\BIBentryALTinterwordstretchfactor\fontdimen3\font minus
  \fontdimen4\font\relax}
\providecommand{\BIBforeignlanguage}[2]{{%
\expandafter\ifx\csname l@#1\endcsname\relax
\typeout{** WARNING: IEEEtran.bst: No hyphenation pattern has been}%
\typeout{** loaded for the language `#1'. Using the pattern for}%
\typeout{** the default language instead.}%
\else
\language=\csname l@#1\endcsname
\fi
#2}}
\providecommand{\BIBdecl}{\relax}
\BIBdecl

\bibitem{thrun2005book}
S.~Thrun, W.~Burgard, and D.~Fox, \emph{Probabilistic robotics}.\hskip 1em plus
  0.5em minus 0.4em\relax MIT press, 2005.

\bibitem{suarez2018can}
F.~Su{\'a}rez-Ruiz, X.~Zhou, and Q.-C. Pham, ``Can robots assemble an ikea
  chair?'' \emph{Science Robotics}, vol.~3, no.~17, p. eaat6385, 2018.

\bibitem{smith1990ARV}
R.~Smith, M.~Self, and P.~Cheeseman, ``Estimating uncertain spatial
  relationships in robotics,'' in \emph{Autonomous robot vehicles}.\hskip 1em
  plus 0.5em minus 0.4em\relax Springer, 1990, pp. 167--193.

\bibitem{durrant1998JRA}
H.~F. Durrant-Whyte, ``Uncertainty geometry in robotics,'' \emph{IEEE Journal
  of Robotics and Automation}, vol.~4, no.~1, pp. 23--31, 1988.

\bibitem{Wang08IJRR}
Y.~Wang and G.~S. Chirikjian, ``Nonparametric second-order theory of error
  propagation on motion groups,'' \emph{International Journal of Robotics
  Research}, vol.~27, no. 11-12, pp. 1258--1273, 2008.

\bibitem{Barfoot14tr}
T.~D. Barfoot and P.~T. Furgale, ``Associating uncertainty with
  three-dimensional poses for use in estimation problems,'' \emph{IEEE
  Transactions on Robotics}, vol.~30, no.~3, pp. 679 -- 693, June 2014.

\bibitem{durrant2006RAM}
H.~Durrant-Whyte and T.~Bailey, ``Simultaneous localization and mapping: part
  i,'' \emph{Robotics \& Automation Magazine, IEEE}, vol.~13, no.~2, pp.
  99--110, 2006.

\bibitem{sallinen2003modelling}
M.~Sallinen, ``Modelling and estimation of spatial relationships in
  sensor-based robot workcells,'' Ph.D. dissertation, University of Oulu (Oulu,
  Finland), 2003.

\bibitem{Su91icra}
S.~Su and C.~Lee, ``Uncertainty manipulation and propagation and verification
  of applicability of actions in assembly tasks,'' in \emph{IEEE International
  Conference on Robotics and Automation}, vol.~3, 1991, pp. 2471--2476.

\bibitem{su1992manipulation}
S.-F. Su and C.~G. Lee, ``Manipulation and propagation of uncertainty and
  verification of applicability of actions in assembly tasks,'' \emph{IEEE
  Transactions on Systems, Man, and Cybernetics}, vol.~22, no.~6, pp.
  1376--1389, 1992.

\bibitem{Taylor1976thesis}
R.~H. Taylor, ``The synthesis of manipulator control programs from task-level
  specifications.'' Ph.D. dissertation, Stanford University, Stanford, CA, USA,
  1976, aAI7707174.

\bibitem{brooks1982IJRR}
R.~A. Brooks, ``Symbolic error analysis and robot planning,'' \emph{The
  International Journal of Robotics Research}, vol.~1, no.~4, pp. 29--78, 1982.

\bibitem{su1992TSMC}
S.-F. Su and C.~G. Lee, ``Manipulation and propagation of uncertainty and
  verification of applicability of actions in assembly tasks,'' \emph{Systems
  Science and Cybernetics, IEEE Transactions on}, vol.~22, no.~6, pp.
  1376--1389, 1992.

\bibitem{brooks1985icra}
R.~A. Brooks, ``Visual map making for a mobile robot,'' in \emph{Robotics and
  Automation, 1985. Proceedings. IEEE International Conference on},
  vol.~2.\hskip 1em plus 0.5em minus 0.4em\relax IEEE, 1985, pp. 824--829.

\bibitem{smith1986jra}
R.~C. Smith and P.~Cheeseman, ``On the representation and estimation of spatial
  uncertainty,'' \emph{IEEE Journal of Robotics and Automation}, vol.~5, no.~4,
  pp. 56--68, 1986.

\bibitem{chirikjian2009vol1}
G.~Chirikjian, \emph{Stochastic Models, Information Theory, and Lie Groups,
  Volume 1: Classical Results and Geometric Methods}.\hskip 1em plus 0.5em
  minus 0.4em\relax Springer Science \& Business Media, 2009, vol.~1.

\bibitem{chirikjian2011vol2}
------, \emph{Stochastic Models, Information Theory, and Lie Groups, Volume 2:
  Analytic Methods and Modern Applications}.\hskip 1em plus 0.5em minus
  0.4em\relax Springer Science \& Business Media, 2011, vol.~2.

\bibitem{knepper2013ikeabot}
R.~A. Knepper, T.~Layton, J.~Romanishin, and D.~Rus, ``Ikeabot: An autonomous
  multi-robot coordinated furniture assembly system,'' in \emph{Robotics and
  Automation (ICRA), 2013 IEEE International Conference on}.\hskip 1em plus
  0.5em minus 0.4em\relax IEEE, 2013, pp. 855--862.

\bibitem{wahrburg2014contact}
A.~Wahrburg, S.~Zeiss, B.~Matthias, and H.~Ding, ``Contact force estimation for
  robotic assembly using motor torques,'' in \emph{Automation Science and
  Engineering (CASE), 2014 IEEE International Conference on}.\hskip 1em plus
  0.5em minus 0.4em\relax IEEE, 2014, pp. 1252--1257.

\bibitem{van2018comparative}
K.~Van~Wyk, M.~Culleton, J.~Falco, and K.~Kelly, ``Comparative peg-in-hole
  testing of a force-based manipulation controlled robotic hand,'' \emph{IEEE
  Transactions on Robotics}, vol.~34, no.~2, pp. 542--549, 2018.

\bibitem{phillips2017planning}
C.~Phillips-Grafflin and D.~Berenson, ``Planning and resilient execution of
  policies for manipulation in contact with actuation uncertainty,''
  \emph{arXiv preprint arXiv:1703.10261}, 2017.

\bibitem{sieverling2017interleaving}
A.~Sieverling, C.~Eppner, F.~Wolff, and O.~Brock, ``Interleaving motion in
  contact and in free space for planning under uncertainty,'' in
  \emph{Intelligent Robots and Systems (IROS), 2017 IEEE/RSJ International
  Conference on}.\hskip 1em plus 0.5em minus 0.4em\relax IEEE, 2017, pp.
  4011--4073.

\bibitem{wirnshofer2018robust}
F.~Wirnshofer, P.~S. Schmitt, W.~Feiten, G.~v. Wichert, and W.~Burgard,
  ``Robust, compliant assembly via optimal belief space planning,'' in
  \emph{2018 IEEE International Conference on Robotics and Automation
  (ICRA)}.\hskip 1em plus 0.5em minus 0.4em\relax IEEE, 2018, pp. 1--5.

\bibitem{platt2010belief}
R.~Platt~Jr, R.~Tedrake, L.~Kaelbling, and T.~Lozano-Perez, ``Belief space
  planning assuming maximum likelihood observations,'' in \emph{Proceedings of
  the Robotics: Science and Systems Conference, 6th}, 2010.

\bibitem{melchior2007particle}
N.~A. Melchior and R.~Simmons, ``Particle rrt for path planning with
  uncertainty,'' in \emph{Robotics and Automation, 2007 IEEE International
  Conference on}.\hskip 1em plus 0.5em minus 0.4em\relax IEEE, 2007, pp.
  1617--1624.

\bibitem{hauser2010randomized}
K.~Hauser, ``Randomized belief-space replanning in partially-observable
  continuous spaces,'' in \emph{Algorithmic Foundations of Robotics IX}.\hskip
  1em plus 0.5em minus 0.4em\relax Springer, 2010, pp. 193--209.

\bibitem{PR97acm}
F.~C. Park and B.~Ravani, ``Smooth invariant interpolation of rotations,''
  \emph{ACM Transactions on Graphics (TOG)}, vol.~16, no.~3, pp. 277--295,
  1997.

\bibitem{huy18tro}
H.~Nguyen and Q.~Pham, ``On the covariance of x in ax=xb,'' \emph{IEEE
  Transactions on Robotics}, vol.~34, no.~6, pp. 1651--1658, 2018.

\bibitem{mason1986mechanics}
M.~T. Mason, ``Mechanics and planning of manipulator pushing operations,''
  \emph{The International Journal of Robotics Research}, vol.~5, no.~3, pp.
  53--71, 1986.

\bibitem{goyal1991planar}
S.~Goyal, A.~Ruina, and J.~Papadopoulos, ``Planar sliding with dry friction
  part 1. limit surface and moment function,'' \emph{Wear (Amsterdam,
  Netherlands)}, vol. 143, no.~2, pp. 307--330, 1991.

\bibitem{petrovskaya2011global}
A.~Petrovskaya and O.~Khatib, ``Global localization of objects via touch,''
  \emph{IEEE Transactions on Robotics}, vol.~27, no.~3, pp. 569--585, 2011.

\bibitem{nguyen2017touch}
H.~Nguyen and Q.-C. Pham, ``Touch-based object localization in cluttered
  environments,'' \emph{arXiv preprint arXiv:1709.09317}, 2017.

\bibitem{zhou2017probabilistic}
J.~Zhou, R.~Paolini, A.~M. Johnson, J.~A. Bagnell, and M.~T. Mason, ``A
  probabilistic planning framework for planar grasping under uncertainty,''
  \emph{IEEE Robotics and Automation Letters}, vol.~2, no.~4, pp. 2111--2118,
  2017.

\end{thebibliography}
\end{document}